\documentclass{article}
\usepackage{spconf,amsmath,graphicx}
\usepackage{url}

\title{Optimal Copula Transport for Clustering Multivariate Time Series}
\name{Gautier Marti$^{\star \dagger}$ \qquad Frank Nielsen$^{\dagger}$ \qquad Philippe Donnat$^{\star}$}

\address{$^{\star}$ Hellebore Capital Management \qquad
    $^{\dagger}$ Ecole Polytechnique}

\begin{document}
%
\maketitle
\begin{abstract}
This paper presents a new methodology for clustering multivariate time series leveraging optimal transport between copulas. Copulas are used to encode both (i) intra-dependence of a multivariate time series, and (ii) inter-dependence between two time series. Then, optimal copula transport allows us to define two distances between multivariate time series: (i) one for measuring intra-dependence dissimilarity, (ii) another one for measuring inter-dependence dissimilarity based on a new multivariate dependence coefficient which is robust to noise, deterministic, and which can target specified dependencies.
\end{abstract}
\section{Introduction}
\label{sec:intro}
Clustering is the task of grouping a set of objects in such a way that objects in the same group, also called cluster, are more similar to each other than those in different groups. This primitive in unsupervised machine learning is known to be hard to formalize and hard to solve. For practitioners, the proper choice of a pair-wise similarity measure, features or representation, normalizations, and number of clusters is a supplementary burden: it is most often task and goal dependent. Time series, sequences of data points or ordered sets of random variables, add complexity to the clustering task being dynamical objects. In the survey \cite{liao2005clustering}, the author classifies time series clustering into three main approaches: working (i) on raw data (e.g., time-frequency \cite{shumway2003time}), (ii) on features (e.g., wavelets, SAX \cite{lin2007experiencing}), (iii) on models (e.g., ARIMA time series \cite{kalpakis2001distance}). Regardless of the method chosen, dependence between time series (usually measured with Pearson linear correlation) is a major information to study. This is notably the case for fMRI, EEG and financial time series. Obviously, dependence does not amount for the whole information in a set of time series.
For example, in the specific case of $N$ time series whose observed values are drawn from $T$ independent and identically distributed random variables, one should take into account all the available information in these $N$ time series, i.e. dependence between them and the $N$ marginal distributions, in order to design a proper distance for clustering \cite{donnat2016toward}.    
Many of the time series datasets which can be found in the literature consist in $N$ real-valued variables observed $T$ times, while in this work we will focus on $N \times d \times T$ time series datasets, i.e. $N$ vector-valued variables in $\mathbf{R}^d$ observed $T$ times. For instance, $N$ horses can be monitored through time using $d$ sensors placed on their body and limbs. Clustering these multivariate time series using dependence information only is already challenging: (i) dependence information can be found at two levels, intra-dependence between the $d$ time series $(x_1^{(i)},\ldots,x_d^{(i)})$, $1 \leq i \leq N$, and inter-dependence between the $N$ time series $(X_1,\ldots,X_N)$; (ii) efficient multivariate dependence measures are required.
Back to the previous example, intra-dependence between the $d$ time series quantifies how the $d$ sensors jointly move and thus may help to cluster horses based on their gaits (e.g., walk, trot, canter, gallop, pace) while inter-dependence between the $N$ time series quantifies how the $N$ horses jointly move and thus may help to cluster horses based on their trajectories.
Recently, several new dependence coefficients between random variables have been proposed in the literature (\cite{poczos2012copula}, \cite{lopez2013randomized}, \cite{ding2013copula}, \cite{kinney2014equitability}) 
demonstrating the interest and the difficulty of obtaining such measures. However, for the clustering task of multivariate time series, most of them are inappropriate: (i) some are not multivariate measures, (ii) some are not robust, i.e. estimate dependence may strongly vary from one estimation to another and may yield erroneously high dependence estimate between independent variables as it is in the case with the Randomized Dependence Coefficient (RDC) \cite{lopez2013randomized} as noticed in \cite{ding2013copula}, (iii) they aim to capture a wide range of dependence equitably \cite{kinney2014equitability} and thus are not application-oriented, i.e. one cannot specify the dependencies we want to focus on and ignore the others. Consequently, shortcoming (ii) leads to spurious clusters, and (iii) to ill-suited clusters for specific tasks besides increasing the risk of capturing spurious dependence, e.g., the Hirschfeld-Gebelein-Renyi Maximum Correlation Coefficient  equals 1 too often \cite{ding2013copula}.
In this work, we will therefore propose a new multivariate dependence measure which was motivated by applications (clustering credit default swaps based on their noisy term structure time series \cite{marti2015proposal}), and the need of robustness on finite noisy samples. 
Our dependence measure leveraging statistical robustness of empirical copulas and optimal transport achieves the best results on the benchmark datasets, yet the experiment is biased in our favor since we specify to our coefficient the dependence we look for (by definition) unlike the other dependence measures.

\subsection*{Contributions}

In this article, we will introduce (i) a method to compare intra-dependence between two multivariate time series, (ii) a dependence coefficient to evaluate the inter-dependence between two such time series, (iii) a method that allows to specify the dependencies our coefficient should measure.
The novel dependence coefficient proposed is benchmarked on experiments \cite{simon2014comment} based on R code from \cite{lopez2013randomized}\footnote{\url{https://github.com/lopezpaz/randomized_dependence_coefficient}}. Tutorial, implementation and illustrations are available at \url{www.datagrapple.com/Tech}.

\section{Related work}
\label{sec:format}

Clustering multivariate time series (MTS) datasets has been much less explored than clustering univariate time series \cite{dasu2005grouping} despite their ubiquity in fields such as motion recognition (e.g., gaits), medicine (e.g., EEG, fMRI) and finance (e.g., fixed-income securities yield curves or term structures).
A general trend for clustering $N$ multivariate time series is to consider them as $N$ datasets of dimension $d \times T$, then in order to obtain a clustering of the $N$ datasets, one leverages a similarity measure between two such $d \times T$ MTS datasets among Euclidean Distance, Dynamic Time Warping, Weighted Sum SVD, PCA similarity factor and other PCA-based similarity measures \cite{yang2004pca} before running a standard algorithm such as $k$-means.
In \cite{singhal2002clustering}, the authors improve on the PCA-based methodology by adding another similarity factor: a Mahalanobis distance similarity factor which discriminates between two datasets that may have similar spatial orientation (similar principal components) but are located far apart. Authors finally combine orientation (PCA) and location (Mahalanobis distance) with a convex combination to feed a $k$-means algorithm leveraging the resulting dissimilarities. Paving another way for research in \cite{dasu2005grouping}, the authors map each individual time series to a fixed-length vector of non-parametric statistical summaries before applying $k$-means on this feature space. In this work, we focus instead on dependence which is not yet well understood in the multivariate setting (e.g., many different definitions of mutual information, the copula construction breaks down for non-overlapping multivariate marginals \cite{genest1995impossibilite}). To alleviate the former shortcoming, we propose to study separately intra-dependence and inter-dependence. In line with the related research, we focus on defining proper distances between the multivariate time series 
rather than elaborating on the clustering algorithm.

\section{Clustering intra-dependence}

We refer to the dependence between the $d$ univariate time series of a $d$-variate time series as intra-dependence. We present the mathematical tools to capture this intra-dependence and how to compare it between two $d$-variate time series in order to perform a clustering based on this information.

\subsection{The Copula Transform}

Since ``the study of copulas and their applications in statistics is a rather modern phenomenon" and ``despite overlapping goals of multivariate modeling and dependence identification,
until recently the fields of machine learning in general [\ldots] have been ignorant of the framework of copulas" \cite{elidan2013copulas}, we recall in this section the basic definitions and results of Copula Theory required for clustering with optimal copula transport.

\vspace{0.2cm}

\textbf{Definition.} \textit{The Copula Transform.} Let $X = (X_1,\ldots,X_d)$ be a random vector with continuous marginal cumulative distribution functions (cdfs) $P_i$, $1 \leq i \leq d$. The random vector $U = (U_1,\ldots,U_d) := P(X) = (P_1(X_1),\ldots,P_d(X_d))$ is known as the copula transform. $U_i$, $1 \leq i \leq d$, are uniformly distributed on $[0,1]$ (the probability integral transform): 
for $P_i$ the cdf of $X_i$, we have $x = P_{i}({P_{i}}^{-1}(x)) = \mathrm{Pr}(X_i \leq {P_{i}}^{-1}(x)) = \mathrm{Pr}(P_{i}(X_i) \leq x)$, thus $P_{i}(X_i) \sim \mathcal{U}[0,1]$.

\vspace{0.2cm}

\textbf{Theorem.} \textit{Sklar's Theorem \cite{sklar1959fonctions}.} For any random vector $X = (X_1,\ldots,X_d)$ having continuous marginal cdfs $P_i$, $1 \leq i \leq d$, its joint cumulative distribution $P$ is uniquely expressed as
$$P(X_1,\ldots,X_d) = C(P_1(X_1),\ldots,P_d(X_d)),$$
where $C$, the multivariate distribution of uniform marginals, is known as the copula of $X$.

\vspace{0.2cm}

Copulas are central for studying the dependence between random variables: their uniform marginals jointly encode all the dependence. One can observe that in most cases, we do not know a priori the margins $P_i$, $1 \leq i \leq d$, for applying the copula transform on $(X_1,\ldots,X_d)$.
\cite{deheuvels1979fonction} has introduced a practical estimator for the uniform margins and the underlying copula, the empirical copula transform.

\vspace{0.2cm}

\textbf{Definition.} \textit{The Empirical Copula Transform \cite{deheuvels1979fonction}.} 

Let $(X_1^t,\ldots,X_d^t)$, $t = 1,\ldots,T$, be $T$ observations from a random vector $(X_1,\ldots,X_d)$ with continuous margins.
Since one cannot directly obtain the corresponding copula observations $(U_1^t,\ldots,U_d^t) = (P_1(X_1^t),\ldots,P_d(X_d^t))$, where $t = 1,\ldots,T$, without knowing a priori $(P_1,\ldots,P_d)$, one can instead estimate the $d$ empirical margins
$P_i^T(x) = \frac{1}{T} \sum_{t=1}^T \textbf{1}(X_i^t \leq x)$, $1 \leq i \leq d$, to obtain the $T$ empirical observations
$(\tilde{U_1^t},\ldots,\tilde{U_d^t}) = (P_1^T(X_1^t),\ldots,P_d^T(X_d^t))$. Equivalently, since $\tilde{U_i^t} = R_i^t / T$, $R_i^t$ being the rank of observation $X_i^t$, the empirical copula transform can be considered as the normalized rank transform.

%
%

\vspace{0.2cm}

\begin{figure}
   \begin{minipage}[c]{.46\linewidth}
      \includegraphics[width=\textwidth]{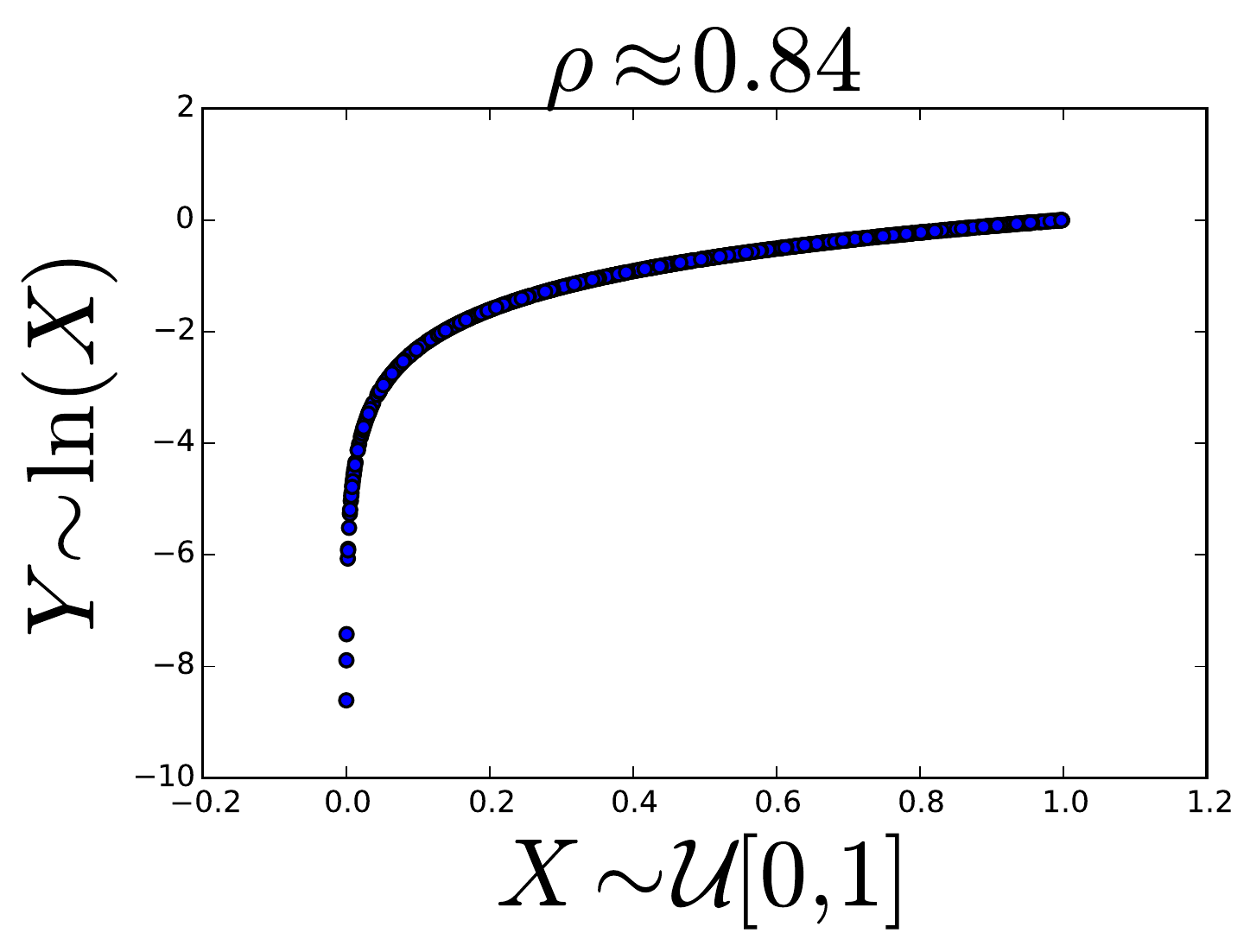}
   \end{minipage} \hfill
   \begin{minipage}[c]{.46\linewidth}
      \includegraphics[width=\textwidth]{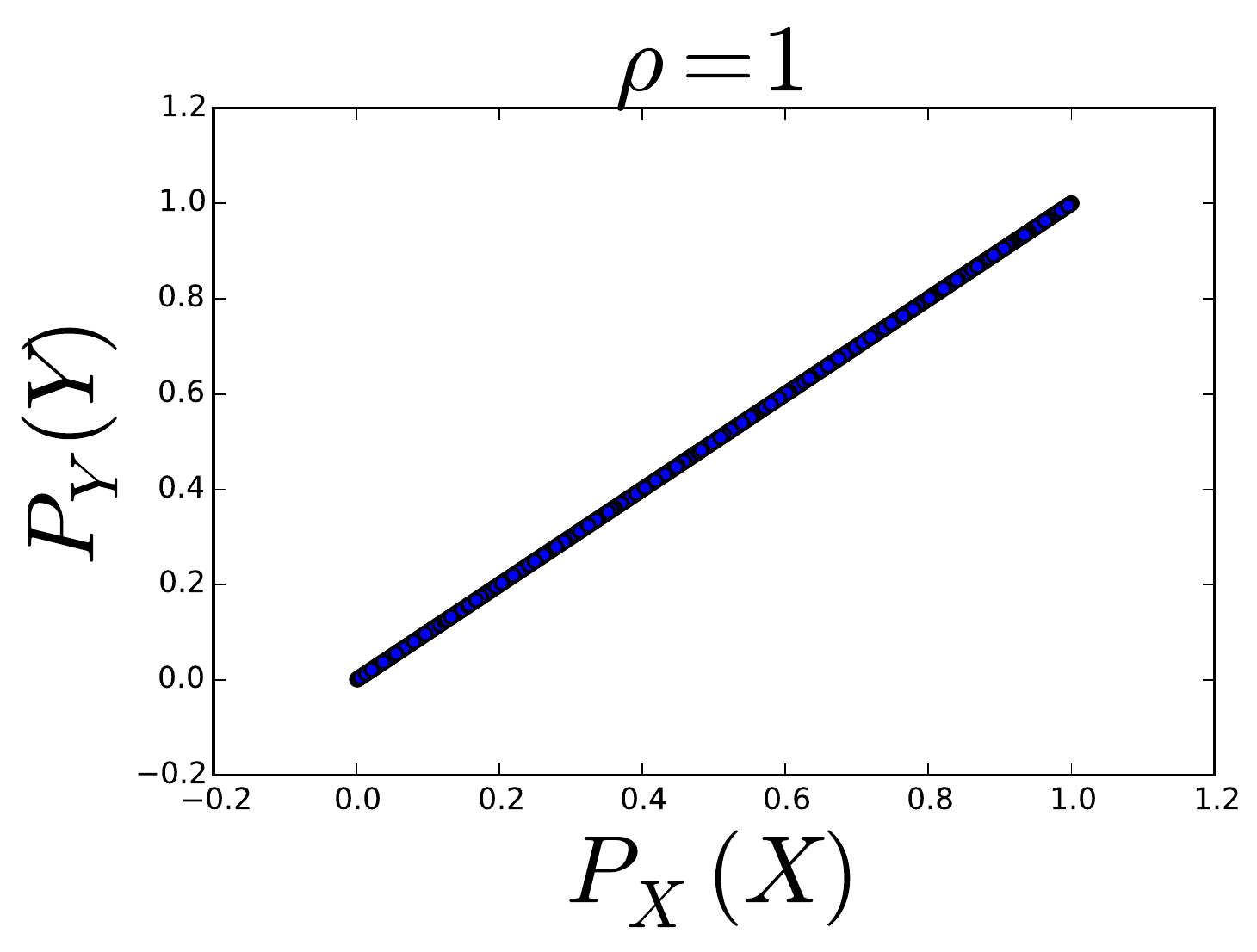}
   \end{minipage}
	\caption{The copula transform invariance property to strictly increasing transformation: Let $X \sim \mathcal{U}[0,1]$ and $Y \sim \ln(X)$. Pearson correlation cannot retrieve the perfect deterministic dependence on raw data (left panel) but it can on the copula transform (right panel).}\label{fig:invariance_incr}
\end{figure}

Few remarks: the empirical copula transform is (i) easy and fast to compute, i.e. sorting $D$ arrays of length $T$, $\mathcal{O}(D T \log T)$; (ii) consistent and converges fast to the underlying copula \cite{deheuvels1981asymptotic}, \cite{poczos2012copula}.
Authors leverage the empirical copula transform for several purposes: \cite{poczos2012copula} benefit from its invariance to strictly increasing transformation of $X_i$ variables (Fig.~\ref{fig:invariance_incr}) for improving feature selection, \cite{lopez2013randomized} to obtain a dependence coefficient invariant with
respect to marginal distribution transformations, and \cite{donnat2016toward} to study separately dependence and margins for clustering.

\subsection{Optimal Transport between Copulas}

Optimal transport is an old problem in applied mathematics anchored in Gaspard Monge seminal treatise which has received a renewed attention from both the pure and applied mathematics communities.
Its recent theoretical development are detailed in \cite{villani2008optimal}.
Meantime, applications have been found in several domains such as mathematical economics \cite{kantorovich1942translocation}, image retrieval \cite{rubner2000earth}, image recoloring \cite{ferradans2013regularized} and univariate empirical probability distributions clustering \cite{henderson2015ep}.

\vspace{0.2cm}

Solving an optimal transport problem amounts to find the optimal transportation and allocation of resources: for instance, finding the best mapping between $n$ factories and $m$ retail stores given the cost of shipment, or the best way to turn piles of dirt into another minimizing the work defined by the amount of dirt moved times the distance by which it is moved.
The last example actually motivated the definition of a distance between two multi-dimensional distributions called the Earth Mover's Distance (EMD) \cite{rubner2000earth} which has been found to be the discrete version of the first Wasserstein distance $W_1(\mu,\nu) := \inf_{\gamma \in \Gamma(\mu,\nu)} \int_{M \times M} d(x,y) \mathrm{d}\gamma(x,y) $ whose definition is strongly related to the optimal transport in its Kantorovich's formulation $\inf \{ \int_{X \times Y} c(x,y)  \mathrm{d}\gamma(x,y) ~| ~ \gamma \in \Gamma(\mu,\nu) \}$. Since copulas encode all the dependencies with their uniform marginals, we leverage them to quantify intra-dependence similarity between two $d$-variate random variables, i.e. how similar the dependence between their $d$ coordinates is.

\vspace{0.15cm}

\textbf{Definition.} \textit{Earth Mover's Distance between two copulas.}

Let $\tilde{U_1}, \tilde{U_2} \in [0,1]^d$ be the empirical copula transforms of data $X_1, X_2 \in \mathbf{R}^{d \times T}$.
We estimate the two empirical copula densities using histograms $h_1$ and $h_2$ that are converted into signatures $s_1 = \{ (p_i, w_{p_i})_{i=1}^n \}$ and $s_2 = \{ (q_i, w_{q_i})_{i=1}^n \}$, where $p_i, q_i \in [0,1]^d$ are the central positions of bins in $h_1$, $h_2$ respectively, and $w_{p_i}, w_{q_i}$ are equal to the corresponding bin frequencies.
We define the intra-dependence distance $D_{\mathrm{intra}}(X_1,X_2) := \mathrm{EMD(s_1,s_2)}$, where \cite{rubner2000earth}

\begin{equation}
\begin{aligned}
& \underset{}{\mathrm{EMD}(s_1,s_2) :=}
& & \min_{f} \sum_{1 \leq i,j \leq n} \|p_i - q_j\| f_{ij} \\
& \text{subject to}
& & f_{ij} \geq 0, \; 1 \leq i,j \leq n, \\
&&& \sum_{j=1}^n f_{ij} \leq w_{p_i}, \; 1 \leq i \leq n, \\
&&& \sum_{i=1}^n f_{ij} \leq w_{q_j}, \; 1 \leq j \leq n, \\
&&& \sum_{i=1}^n \sum_{j=1}^n f_{ij} = 1.
\end{aligned}
\end{equation}

\vspace{0.2cm}

From a practical point of view, this distance is robust to the binning process (i.e. small misalignements of corresponding frequencies yield to a slightly larger distance) unlike the standard bin-by-bin distances (small misalignements yield to a much larger distance) such as Kullback-Leibler divergence for instance.
However, its main drawback is the computational complexity. Finding the optimal transport between the $n$ Diracs is an instance of the assignement problem. This fundamental combinatorial optimization problem is equivalent to the problem of finding a minimum weight matching in a complete weighted bipartite graph $K_{n,n}$ which can be solved by the Hungarian algorithm in $\mathcal{O}(n^3)$. 

\begin{figure}[htp]
\begin{center}
\includegraphics[width=\linewidth]{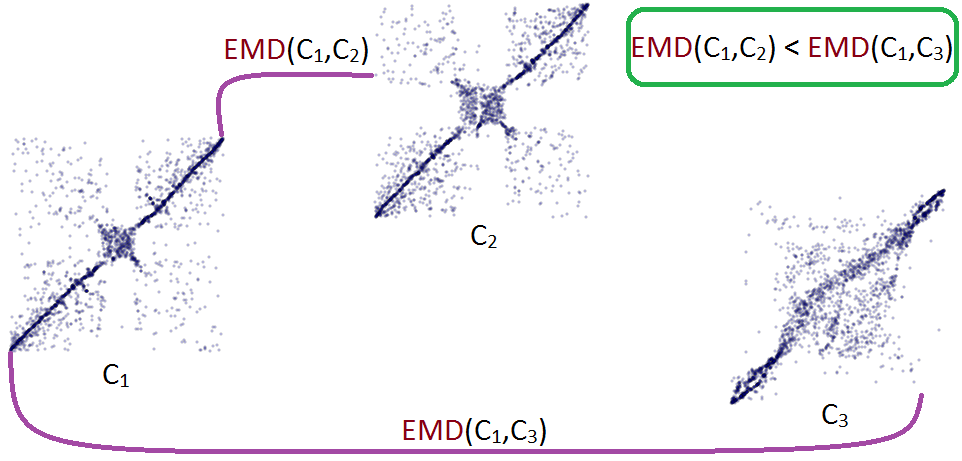}
\caption{Optimal Copula Transport for measuring intra-dependence similarity of two MTS; $C_1$, $C_2$, $C_3$ are empirical copulas of data $X_1, X_2, X_3 \in \mathbf{R}^{2 \times T}$ respectively. According to the EMD, the dependence between the two coordinates of $X_1$ is much similar to those of $X_2$ than those of $X_3$. 
}\label{fig:copula_transport_intra}
\end{center}
\end{figure}

In Fig.~\ref{fig:copula_transport_intra}, we display three bivariate empirical copulas computed on real data (these copulas encode the dependence between two maturities of the credit default swaps for three different entities). We can notice strong positive dependence (expressed by the diagonals), yet these copulas exhibit distinct dependence patterns which may not be taken into account by correlation coefficients (cf. the tutorial at \url{www.datagrapple.com/Tech} for an application of this methodology to credit default swaps time series). 

\section{Clustering inter-dependence}

\begin{figure}[htp]
\begin{center}
\includegraphics[width=\linewidth]{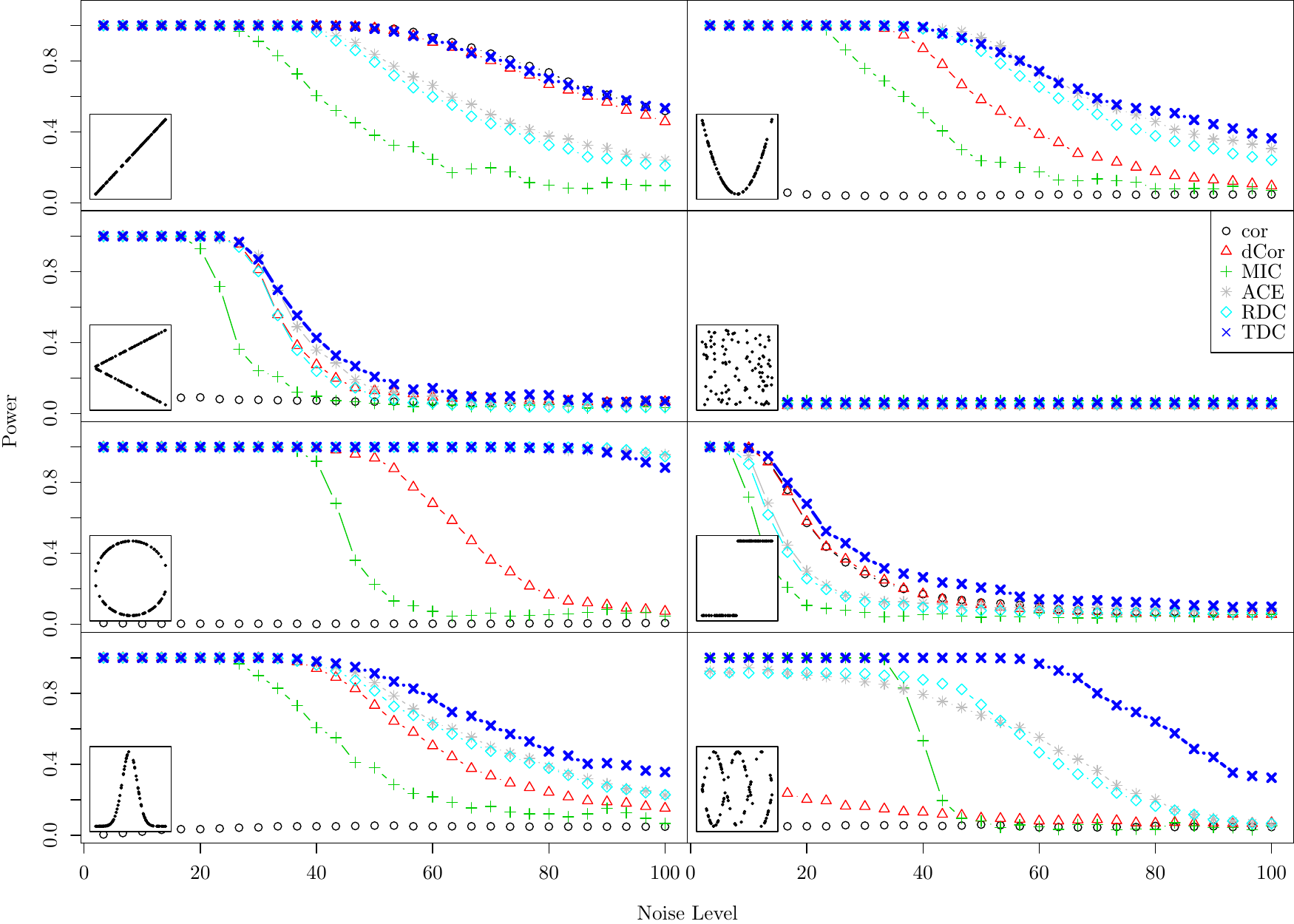}
\caption{Experiments based on \cite{simon2014comment} and \cite{lopez2013randomized};
Dependence estimators power as a function of the noise for several deterministic patterns + noise. Their power is the percentage of times that they are able to distinguish between dependent and independent samples. TDC (dark blue curves) can deal with complex dependence patterns in presence of noise if they are specified in its targets.}\label{fig:power_estimate}
\end{center}
\end{figure}

We refer to the dependence between two $d$-variate time series as inter-dependence. In this section, we present our inter-dependence measure and compare it in Fig.~\ref{fig:power_estimate} with the most popular coefficients in the literature (ACE, 
dCor, 
MIC, 
RDC). 

\begin{figure}[htp]
\begin{center}
\includegraphics[width=\linewidth]{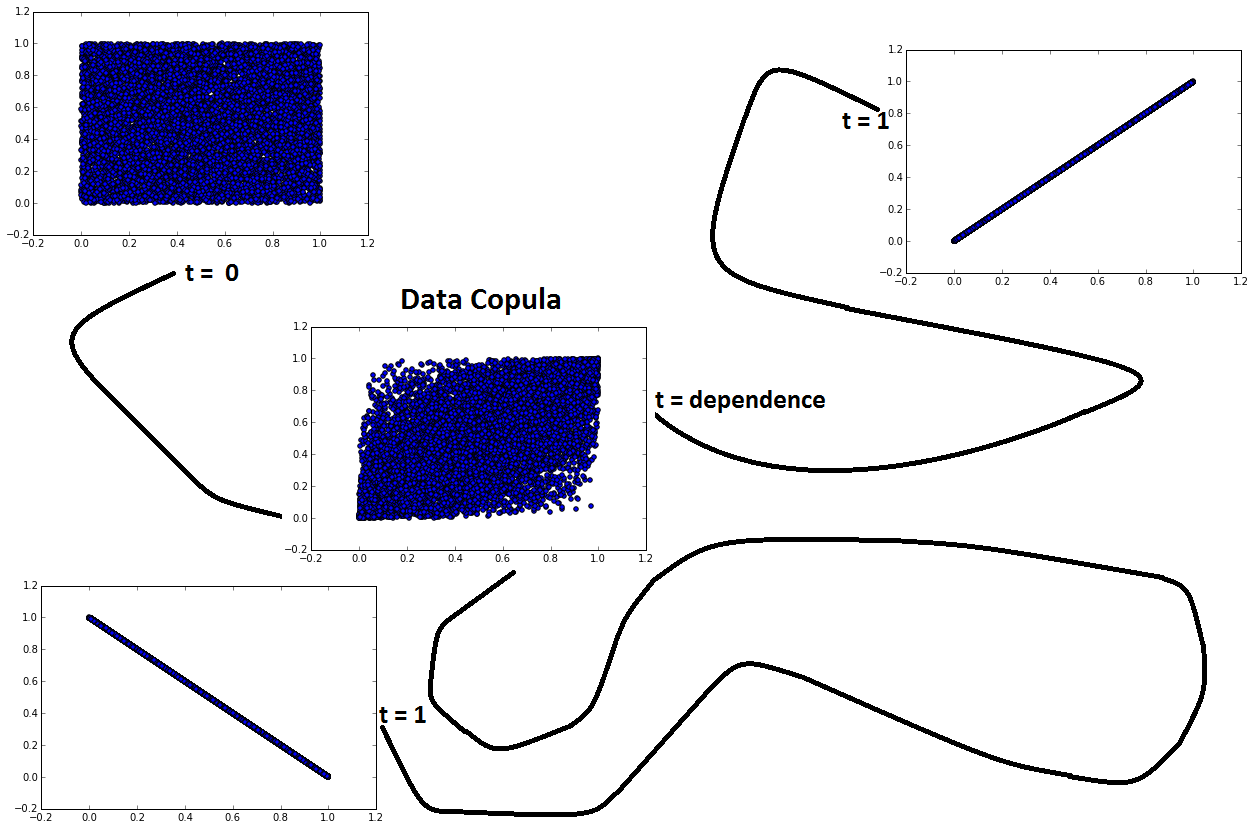}
\caption{Optimal Copula Transport for estimating dependence between random variables; Dependence can be seen as the relative distance between the independence copula and one or more target dependence copulas. In this picture, the target dependencies are ``perfect dependence" and ``perfect anti-dependence". The empirical copula (Data Copula) was built from positively correlated Gaussians, and thus is nearer to the ``perfect dependence" copula (top right corner) than to the ``perfect anti-dependence" copula (bottom left corner).}\label{fig:copula_transport}
\end{center}
\end{figure}

Let $Z^T = (X_1^T,\ldots,X_d^T,Y_1^T,\ldots,Y_d^T) \in \mathbf{R}^{2d \times T}$ be $T$ stacked observations from the random vectors $X = (X_1,\ldots,X_d)$ and $Y = (Y_1,\ldots,Y_d)$.
Informally, in order to estimate dependence between $X$ and $Y$, we use the relative position of the empirical copula $\tilde{C}$ built from data $Z^T$ on the shortest path starting from the independence copula $C_{\mathrm{ind}}$, ending to one of the copulas $\{C_i\}$ encoding the target dependencies, and passing through $\tilde{C}$. This idea is depicted in Fig.~\ref{fig:copula_transport} and benchmarked in Fig.~\ref{fig:power_estimate}.

\vspace{0.2cm}

\textbf{Definition.} \textit{Target Dependencies Coefficient using Transport to Dependence Copulas.}
The Target Dependencies Coefficient using Transport to Dependence Copulas (TDC) is defined as
$$\mathrm{TDC}(X^T,Y^T) := \frac{\mathrm{EMD}(C_{\mathrm{ind}},\tilde{C})}{\mathrm{EMD}(C_{\mathrm{ind}},\tilde{C}) + \min_i \mathrm{EMD}(\tilde{C},C_i)}.$$
For $\tilde{C} = C_\mathrm{ind}$, TDC = 0, for $\tilde{C} \in \{C_i\}$, TDC = 1, otherwise TDC quantifies the relative nearness to independence or to the specified dependencies. Bonus: practitioners can discover which specified dependence is ``activated", i.e. which of their hypotheses about the dependence between $X$ and $Y$ seems the most likely (qualitative) and how strong (quantitative).

\section{Discussion}

The proposed methodology presents several \textbf{benefits}: non-parametric, robust and deterministic (optimal transport), accurate and generic representation of dependence (empirical copulas). Yet, it has also some scalability \textbf{drawbacks}: (i) in dimension, non-parametric estimations of density suffer from the curse of dimensionality (ii) in time, EMD is costly to compute (but, there exist methods for speeding up computations \cite{cuturi2013sinkhorn}). To alleviate drawback (i), parametric methods may be a solution, we consider optimal copula transport in statistical manifolds for further research.

\bibliographystyle{IEEEbib}
\bibliography{refs}

\end{document}